\documentclass[sigconf]{acmart}
\usepackage{multirow} 
\usepackage{rotating}
\usepackage{float}

\AtBeginDocument{%
  }

\setcopyright{acmlicensed}
\copyrightyear{2018}
\acmYear{2018}
\acmDOI{XXXXXXX.XXXXXXX}

\acmConference[Conference acronym 'XX]{Make sure to enter the correct
  conference title from your rights confirmation emai}{June 03--05,
  2018}{Woodstock, NY}
\acmISBN{978-1-4503-XXXX-X/18/06}

\begin{document}
\title{ConvTimeNet: A Deep Hierarchical Fully Convolutional Model for Multivariate Time Series Analysis}

\author{Mingyue Cheng\textsuperscript{1}, Jiqian Yang\textsuperscript{1}, Tingyue Pan\textsuperscript{1}, Qi Liu\textsuperscript{1}, Zhi Li\textsuperscript{2}, Shijin Wang\textsuperscript{3}}
\affiliation{
  \institution{\textsuperscript{1}State Key Laboratory of Cognitive Intelligence, University of Science and Technology of China, Hefei, China \\ \textsuperscript{2}Shenzhen International Graduate School, Tsinghua University, Shenzhen, China\\ \textsuperscript{3}Artificial Intelligence Research Institute, iFLYTEK Co., Ltd, Hefei, China}
\city{ }
\country{ }
}

\email{{yangjq,pty12345}@mail.ustc.edu.cn,  {mycheng,qiliuql}@ustc.edu.cn, zhilizl@sz.tsinghua.edu.cn,	sjwang3@iflytek.com}
\begin{abstract}
Designing effective models for learning time series representations is foundational for time series analysis.
Many previous works have explored time series representation modeling approaches and have made progress in this area. Despite their effectiveness, they lack adaptive perception of local patterns in temporally dependent basic units and fail to capture the multi-scale dependency among these units. Instead of relying on prevalent methods centered around self-attention mechanisms, we propose ConvTimeNet, a hierarchical pure convolutional model designed for time series analysis. ConvTimeNet introduces a deformable patch layer that adaptively perceives local patterns of temporally dependent basic units in a data-driven manner. Based on the extracted local patterns, hierarchical pure convolutional blocks are designed to capture dependency relationships among the representations of basic units at different scales. Moreover, a large kernel mechanism is employed to ensure that convolutional blocks can be deeply stacked, thereby achieving a larger receptive field. In this way, local patterns and their multi-scale dependencies can be effectively modeled within a single model. Extensive experiments comparing a wide range of different types of models demonstrate that pure convolutional models still exhibit strong viability, effectively addressing the aforementioned two challenges and showing superior performance across multiple tasks. The code is available for reproducibility\footnote{https://github.com/Mingyue-Cheng/ConvTimeNet}.
\end{abstract}



\keywords{Time series classification, Deep convolution network}

\maketitle

\section{Introduction}
Time series, sequences of data points arranged in chronological order, are fundamental to various fields due to the valuable insights they provide through analysis and mining~\cite{roberts2013gaussian,kim2021reversible}. In the realm of computer networks, time series analysis plays a pivotal role, especially amid the rapid expansion of the internet infrastructures. The ability to analyze and interpret temporal patterns in network data is essential for numerous applications. For instance, monitoring network traffic over time aids in detecting anomalies, predicting congestion, and ensuring network security through intrusion detection systems. Moreover, time series analysis enables network administrators to proactively address performance issues by forecasting bandwidth utilization and latency fluctuations, thereby enhancing the efficiency and reliability of network operations.

Over the past decades, numerous efforts~\cite{box2015time,liu2021forecast} have been dedicated to this area. Initially, classical statistical methods ARIMA~\cite{zhang2003time} were predominant because of their mathematical rigor and interpretability. However, these approaches often struggled with capturing nonlinear patterns and scaling to large datasets. The advent of deep learning introduced data-driven methods such as Long Short-Term Memory (LSTM) networks~\cite{lou2022mts} and Gated Recurrent Units (GRUs)~\cite{petnehazi2019recurrent}, which modeled sequential data by maintaining hidden states that capture temporal dependencies. Despite advancing the field, these models faced challenges like vanishing gradients and computational inefficiency when handling long sequences.

To address these limitations, the convolutional network~\cite{he2016deep,zheng2014time,middlehurst2023bake} has been adapted for time series analysis, largely due to its inherent properties that strike an excellent balance between computational efficiency and representation quality. Data from past years shows that many representative works~\cite{bagnall2017great} of time series analysis typically employ convolutional networks as the backbone. For instance, TCN~\cite{bai2018empirical} and its variants are widely used in modeling temporal variation dependence for the time series forecasting task. Furthermore, a large number of works (such as InceptionTime~\cite{ismail2020inceptiontime}, MiniRocket~\cite{dempster2021minirocket}, and MCNN~\cite{cui2016multi}) are also proposed by employing convolutional networks to identify informative patterns from given instances in the classification of time series. Recently, Transformer-style networks~\cite{wen2022transformers} have nearly become the dominant role in learning representations of time series, ranging from forecasting~\cite{zhou2021informer,liu2023itransformer} to classification tasks~\cite{liu2021gated}. Additionally, novel approaches like Mamba~\cite{gu2023mamba} and adaptations of Graph Neural Networks (GNNs)~\cite{jiang2023spatio} have emerged, offering alternative perspectives in handling time series data. However, with the advent of Transformer networks~\cite{vaswani2017attention,dosovitskiy2020image}, the role of convolutional architecture in time series analysis appears to be diminishing. 

Despite the effectiveness of these methods, two fundamental challenges remain unaddressed: local temporal pattern extraction and capturing global sequence dependency. For the formal, the current approach is to uniformly segment the time series into patches, treating each patch as a local pattern. However, this uniform segmentation may not adaptively capture the most informative local patterns, potentially overlooking critical features. Besides, solely focusing on these patterns is insufficient for time series analysis. This is because, in time series classification, local patterns serve as the basic units of representation, but without considering the dependencies and relationships among them on a global scale, it is impossible to distinguish between different time series classes. It is necessary to analyze the dependencies among local patterns from a global perspective. Regarding the latter, although numerous methods have been proposed to model global dependency relationships, they often focus solely on these dependencies while neglecting the effective extraction of local patterns.

In response to the aforementioned questions, in this work, we propose a novel convolutional network, dubbed ConvTimeNet, designed as a versatile model for time series analysis that simultaneously addresses both local pattern extraction and capturing multi-scale repensentation dependency within a single model. One of the most distinctive features of ConvTimeNet is its use of pure convolutional operators to learn the representation of given time series data. Crucially, the network not only preserves the advanced properties of Transformer networks but also inherits several inherent strengths of convolutional networks. Specifically, we first adhere to the modern philosophy of neighboring sequence points together to avoid the sparsity semantic dilemma carried by a single numerical value. Notably, we propose to divide the raw time series into sequences of patches in a data-driven manner, which can better preserve the semantics of local regions. Secondly, we design a novel fully convolutional block where deepwise and pointwise convolutions are organized together simultaneously. Particularly, the proposed ConvTimeNet highlights that very deep and hierarchical network architectures are encouraged and very significant in modeling global receptive fields and learning multi-scale representations of given time series instances. To demonstrate the effectiveness and generic capacity of ConvTimeNet, we conduct extensive experiments over a variety of time series analysis tasks, including forecasting and classification. The experimental results show that ConvTimeNet could achieve superior or competitive performance compared to strong baselines, including both advanced Transformer networks and pioneering convolutional models. We hope that the proposed ConvTimeNet can serve as an alternative and useful model for time series analysis.

Main contributions of this paper can be summarized as follows:
\vspace{-0.1in}
\begin{itemize}
  \item We propose ConvTimeNet, a novel fully hierarchical convolutional model that addresses both local pattern extraction and global sequence dependency in time series analysis.
  \item We design a convolutional building block that can be deeply stacked by utilizing depthwise and pointwise convolutions along with a reparameterization mechanism.
  \item We construct extensive experiments to demonstrate that pure convolutional models can outperform strong baselines on multiple tasks, and numerous additional experiments further confirm the effectiveness of our model design.
\end{itemize}
\vspace{-0.1in}
\section{Related Work}
Time series local pattern extraction and capturing its dependency modeling are crucial aspects of time series analysis. Over the years, various approaches have been proposed to model these dependencies, ranging from traditional statistical methods to modern data-driven techniques. Early research predominantly focused on statistical methods such as the ARIMA~\cite{zhang2003time}, which constructs autoregressive models and forecasts using moving averages. ARIMA effectively captures linear relationships within stationary time series data by combining autoregressive terms (AR), differencing operations (I), and moving average terms (MA) to model temporal dependencies. In recent years, deep learning has ushered in a new era of data-driven methods for time series analysis, including those rooted in Convolutional Neural Networks (CNN), Multi-Layer Perceptrons (MLP), and Transformer networks. The CNN-based methods, such as those in \cite{zheng2014time,cui2016multi}, employ sliding convolutional kernels along the temporal dimension to capture temporal dependency. However, these methods have not yielded ideal results in modeling long-range dependencies due to the limited receptive field. On the other hand, MLP-based methods, such as those in \cite{zeng2023transformers,challu2023nhits}, utilize the MLP structure to encode temporal dependencies into the MLP layers. Alternatively, some methods, such as those in \cite{das2023long}, can integrate covariate information into the network.

The Transformer network, renowned for its ability to capture long-range dependencies and cross-variable interactions, is particularly appealing for time series analysis. Consequently, numerous Transformer-based methods have been developed. For instance, Autoformer \cite{wu2021autoformer} utilizes a self-correlation mechanism to capture temporal dependencies. Crossformer \cite{zhang2022crossformer} is another Transformer-based method that introduces a customized dimension segmentation embedding scheme and an explicit cross-variable attention module designed for forecasting tasks. FormerTime \cite{cheng2023formertime} and GPTH \cite{liu2024generative}, on the other hand, employ a hierarchical structure to capture different-scale temporal dependencies and variable dependencies. Meanwhile, the temporal patching operation \cite{nie2022time} has significantly improved the performance of the model. Despite this, the application of convolutional networks in time series analysis has gradually become less common.

Recently, the application of Large Language Models (LLMs) to time series analysis \cite{tan2024language, chang2023llm4ts, cheng2024advancing} has attracted significant attention in the research community. In the Time-LLM~\cite{jin2023time}, the input time series is first tokenized through patching and then aligned with low-dimensional word embeddings using multi-head attention. The aligned outputs, combined with embeddings of descriptive statistical features, are then fed into a frozen, pre-trained language model. The output representations from the language model are flattened and passed through a linear layer to generate the forecast. In this process, the patching employs a uniform segmentation approach.

As modern convolutional techniques continue to evolve, a growing number of convolutional methods are being revisited within the community. For instance, the method \cite{ding2022scaling} proposes RepLKNet, which employs a mechanism of re-parameter to circumvent issues associated with expanding the convolutional kernel, thereby addressing the issue of receptive field limitation. Additionally, the TimesNet, introduced recently in \cite{wu2022timesnet}, considers two-dimensional temporal variations generated by periodicities, designed for general tasks in time series analysis.

\section{A Recipe of Fully Convolutional Network for Time Series Analysis}
\begin{figure*}[ht]
\centering
\includegraphics[width=1.0\textwidth]{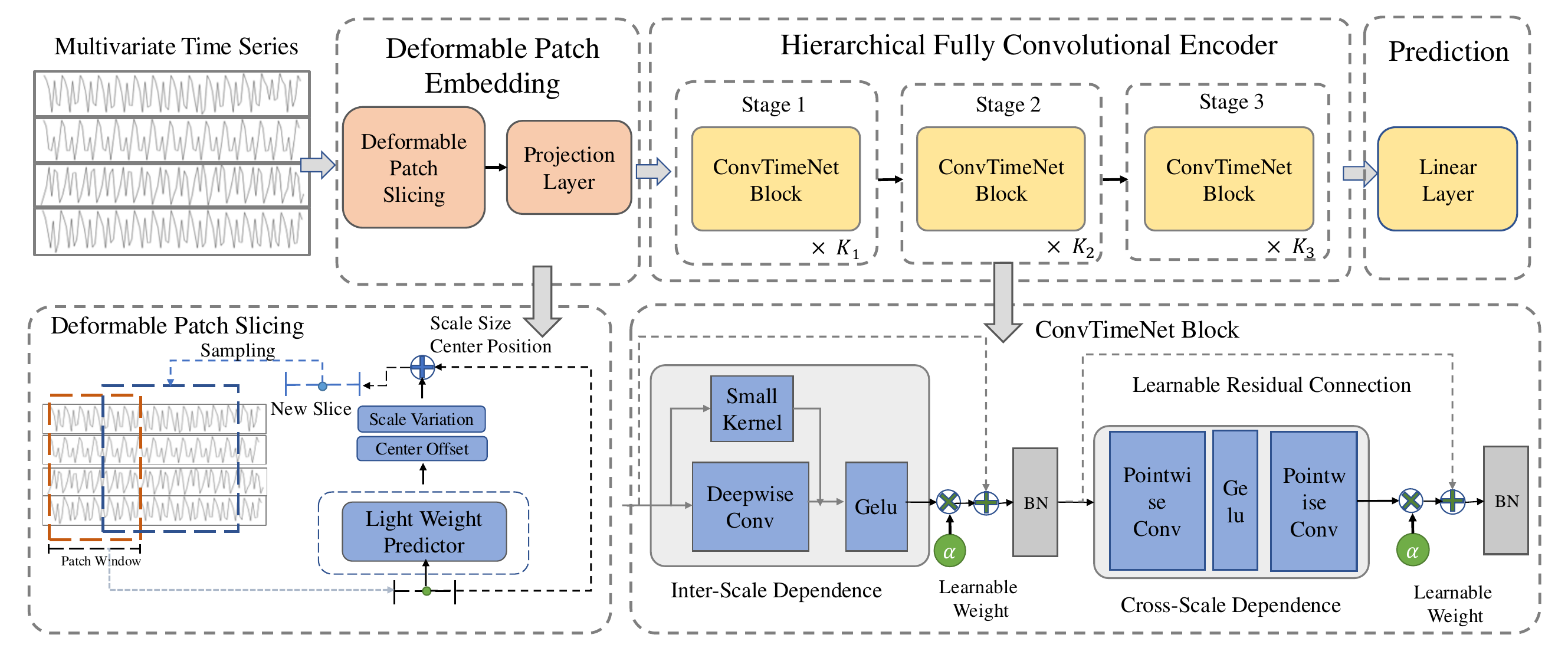}
  \vspace{-0.3in}
  \caption{Illustration of the newly proposed ConvTimeNet.}
  \vspace{-0.1in}
  \label{fig:model}
  \Description{Illustration of the newly proposed ConvTimeNet.}
\end{figure*}
In this section, we begin by providing a brief overview of the architecture of the newly proposed ConvTimeNet. Subsequently, we delve into two key characteristics: deformable patch embedding and fully convolutional blocks. Finally, we demonstrate the strengths of the proposed ConvTimeNet, examining its capabilities from three distinct perspectives.

\subsection{Overall Network Architecture}
As illustrated in Figure~\ref{fig:model}, the forward pass of ConvTimeNet begins with the input of multivariate time series data. To learn the representations of this time series data, we employ a stack of Fully Convolutional Blocks. Leveraging the natural advantages of pure convolution, we vary the sizes of convolutional kernels to deeply model data representations at multiple scales, effectively capturing long-term dependencies. Prior to this, and unlike traditional methods that split patches in a fixed manner, we develop a deformable patch embedding layer. This data-driven patch splitting approach ensures that patterns within each patch are preserved more reasonably, as the selection of time series points for each patch is determined by a predictor layer. Finally, the architecture culminates in a linear layer designed to perform time series mining tasks like forecasting or classification of the given time series data.

\subsection{Adaptively Local Pattern Extraction}
The primary objective of time series analysis is to discern the correlation between data points across different time steps. In most existing studies, preserving raw point-wise time series data as input is nearly a default practice. However, a single-time step does not carry as much semantic meaning as a word in a sentence. Consequently, enhancing the information density of each input unit is crucial for improving the final results. We also note that some recent Transformer studies have begun to attempt to extract local information while analyzing their correlations. A common strategy is to aggregate neighboring time points into sub-series level patches. This process, in contrast to modeling each point-wise input individually, actually groups time points into local regions, thereby coupling their dependencies. While this approach is effective, we argue that the fixed-size patch splitting may compromise semantics, leading to reduced performance. For instance, the same pattern in different time series might have varying sizes, a concept supported by well-known shapelet-based approaches~\cite{ye2009time}. Therefore, strict patch splitting may capture inconsistent information across different instances, likely leading to suboptimal results. To address the dilemma mentioned above, we decided to adopt a philosophy of neighboring time points together, selecting specific time points within each patch in a data-driven manner. This idea of this approach is also consistent with previous work~\cite{dai2017deformable,chen2021dpt}. As illustrated in the lower left of Figure~\ref{fig:model}, we additionally introduced a lightweight predictor to adaptively select time points based on given vanilla features, followed by an embedding projection layer. In this study, we refer to this component as the DePatch module.
 
To achieve this goal, the DePatch module comprises three key steps. Formally, we denote the input time series as $X\in\mathbb{R}^{C\times T}$, where $C$ and $T$ indicate the number of variables (channels) and sequence lengths, respectively. Before performing adaptive patch splitting, we first process the raw time series points into $N = \lfloor{\frac{T-P}{S}}\rfloor + 2$ fixed-size patches, where the shape of each patch can be denoted by $x\in \mathbb{R}^{P \times C}$. Padding is used to ensure a uniform split. For clarity, we omit the superscript ($i$). Here, $P$ denotes the corresponding patch size while $S$ indicates the stride size of the patch window. Subsequently, each patch is assigned three key variables: the center position of this patch $x_c$, center offset $\delta_c$, and scale variation $\delta_p$. It should be noted that $x_c$ is decided by the uniform patch splitting above while $\delta_c$ and $\delta_p$ are obtained by borrowing the following computation process:
 \begin{equation}
     \delta_{c}, \delta_{p} = H(g(x)),
 \end{equation}\noindent where $g(x)$ denotes the feature map obtained from input patch $x$, and $H(\cdot)$ indicates the lightweight predictor function. In practice, the lightweight predictor function can be instantiated by a convolutional-based projection layer. Particularly, extensive experiments over the experimental part show its sensitivity to the final results. Along this line, the final center of patch $x$ and its scale can be computed by 
\noindent
\begin{equation}
\begin{aligned}
    x_{c}^{new} &= x_c + \delta_{c}, 
    P^{new} = P + 2\delta_{p}, \\
    L &= x_{c}^{new} - \frac{P^{new}}{2}, 
    R = x_{c}^{new} + \frac{P^{new}}{2}
\end{aligned}
\end{equation}
\noindent 
where $x_{c}^{new}$ denoted as the new center position of the patch and $P^{new}$ represents the new length of the patch. $L$ and $R$ are denoted as the left and right boundary positions. Based on these two positions, we perform sampling on the original sequence with linear interpolation and obtain a new patch, followed by the projection layer transforms the new patch into the embedding space. With these efforts, the DePatch module allows for the adjustment of each patch’s position and scale based on the input features adaptively, thereby mitigating the semantic destruction caused by hard splitting.

\subsection{Multi-Scale Representation Dependency Modeling}
To learn high-quality data representation, past methods \cite{he2016deep,vaswani2017attention} stacked building blocks, a paradigm that is also very popular in the field of data representation learning. Typically, the Transformer block, as a classical design, has been successfully applied in various fields, and these successful applications have inspired methods in the field of time series. However, the Transformer block also has some drawbacks. On one hand, the traditional Transformer block primarily processes fixed-scale input, which limits its flexibility. On the other hand, the computational complexity of the Transformer block scales quadratically with the length of the input sequence.
 
We propose a fully convolutional block, as shown in the lower right module of Figure~\ref{fig:model}. Specifically, we retain the Transformer block's innovative architecture, incorporating linear and non-linear transformations. The primary distinction lies in our utilization of temporal dependency modeling, achieved through stacking 1D kernel groups instead of the traditional multi-head attention mechanism. Practically, we employ deepwise convolution as a 1D kernel group, as it offers efficient computational efficiency. Meanwhile, we substitute the Feed-Forward Network (FFN) in the Transformer block with a 1x1-sized pointwise convolution to achieve cross-scale dependency modeling. The synergistic combination of these two convolutions effectively completed the comprehensive modeling of the temporal and cross-scale dependency.

To achieve a global receptive field, we increase the depth of the model. We discard the traditional residual connection and instead adopt a learnable residual~\cite{bachlechner2021rezero}, to circumvent the overfitting issue that often arises from an increased model depth. We also verify the effectiveness of the learnable residual mechanism in our experiments. Concurrently, we utilize Batch Normalization~\cite{ioffe2015batch} in a fully convolution block. Furthermore, we employ larger kernels to expand the receptive field. However, in practice, we found that training with large kernels presents a greater challenge. This observation is consistent with recent research finding~\cite{ding2022scaling}. Therefore, we introduce a re-parameterization mechanism to address this issue. 
We divide the entire deep network architecture into multiple distinct stages, each stage is composed of $K$ fully convolutional blocks. Drawing upon this hierarchical architecture, the time series representation across various scales can be effectively extracted. We also explored the impact of different configurations of hierarchical structures on the model's capabilities in the experiment.
 
Concretely, denote $F_{\boldsymbol{\theta}}^{l}(\cdot)$ as the larger kernel branch and $G_{\boldsymbol{\theta}}^{l}(\cdot)$ as the small kernel branch in the $l$-th fully convolutional block. Note that $\boldsymbol{\theta}$ represents the weights of the convolution kernels; they will be updated during gradient backpropagation. The re-parameterization mechanism is defined as follows:
 \begin{equation}
 {Z}_{DW}^{l} = {Gelu}(F_{\boldsymbol{\theta}}^{l}({Z^{l-1}}) + G_{\boldsymbol{\theta}}^{l}({Z^{l-1}})),
 \end{equation}\noindent where $Z^{l-1}$denotes the output of $l-1$ the convolutional block, $Z_{DW}^{l}\in \mathcal{\mathbb{R}}^{D \times M}$ represents the output deepwise convolution layer in the $l$-th fully convolutional block. $D$ denoted as the number of hidden dimensions and $M$ represents the new sequence length, which is equal to the number of patches $N$. During the inference phase in ConvTimeNet, a sophisticated technique is employed where the weights of the small kernel are zero-padded and then combined with those of the larger kernel to form a unified convolutional kernel. Denote $\alpha$ as the learnable weights for the residual connections in the deepwise convolution layer, typically initialized with $0$.
 \begin{equation}
     \hat{Z}_{DW}^{l} = Z^{l-1} + \alpha \times Z_{DW}^{l},
 \end{equation}\noindent where $\hat{Z}_{DW}^{l}$ denote the output of deepwise convolution in $l$-th building block. To sum up, by utilizing fully convolutional blocks, we can effectively model cross-dependencies within a single block. The design of hierarchical and large kernels for ConvTimeNet facilitates the representation learning of varying scales and the creation of a global receptive field.
 
\subsection{Analysis of the ConvTimeNet}
ConvTimeNet enhances the semantic representation of the input data using the deformable patch module, which adaptively extracts local temporal patterns in a data-driven manner. This module directly tackles the issue of local pattern extraction by allowing the model to focus on the most informative segments of the time series rather than relying on uniform segmentation methods that may overlook critical local features. By capturing the nuanced variations within the data, ConvTimeNet preserves essential local semantics crucial for accurate representation learning.

Furthermore, the fully convolutional block in ConvTimeNet is designed with a combination of depthwise and pointwise convolutions, along with a reparameterization mechanism. This architecture enables the model to efficiently capture multi-scale pattern representation dependencies. The use of large convolutional kernels within a hierarchical framework allows ConvTimeNet to attain a global receptive field without the computational complexity associated with self-attention mechanisms in Transformer networks.

By unifying the solutions to both local pattern extraction and global sequence dependency capturing within a single model, ConvTimeNet provides a comprehensive framework that addresses these two fundamental challenges in time series analysis. This unification is significant because it allows ConvTimeNet to simultaneously extract informative local features through the deformable patch embedding and capture multi-scale representation dependencies using the fully convolutional block. By bridging the gap between traditional convolutional networks and Transformer networks, ConvTimeNet achieves superior performance, offering a holistic approach that previous models have not accomplished within a single architecture.
\vspace{-0.1in}
\section{Experiments}
\subsection{Experimental Setup}
\paragraph{Time Series Classification.}
For the task of time series classification, our experiments are carried out on a selection of $10$ representative multivariate time series datasets from the renowned UEA archive, which is consistent with recent works~\cite{liu2021gated,wu2022timesnet,cheng2023formertime}. To demonstrate the effectiveness of our ConvTimeNet in the classification of time series, we chose the following strong competitive baselines, including self-attention based methods: FormerTime~\cite{cheng2023formertime}, TST~\cite{zerveas2021transformer}, Convolutional based: TimesNet~\cite{wu2022timesnet}, MiniRocket~\cite{dempster2021minirocket}, TCN~\cite{bai2018empirical}, MCDCNN~\cite{zheng2014time}. In addition, MLP is also employed as our baseline. We do not choose well-known classical distance-based~\cite{middlehurst2021hive} and shapelet-based~\cite{ye2009time,he2023shapewordnet} approaches due to the concern of computation complexity. For the classification task, the model optimization utilizes ADAM as well. The number of training epochs is fixed at 200. All the baselines reproduced are implemented based on the configurations outlined in the original paper or the official code.
\vspace{-0.1in}
\paragraph{Time Series Forecasting.}
For the task of time series forecasting, we extensively conduct experimental evaluation over $9$ well-known public datasets in our experiments, including ETT (consists of $4$ subsets, i.e.,  ETTh1, ETTh2, ETTm1, ETTm2), Electricity, Exchange, Traffic, Weather, and Illness. To demonstrate the effectiveness of our ConvTimeNet in the prediction of time series, the following advanced approaches are selected as competitive baselines, including convolutional based, such as TimesNet\cite{wu2022timesnet}, models rooted in Multilayer Perceptrons (MLPs), like N-Hits\cite{challu2023nhits} and DLinear\cite{zeng2023transformers}, and those based on Transformers, including iTransformer\cite{liu2023itransformer}, Crossformer\cite{zhang2022crossformer}. Furthermore, we have scrutinized top-tier models specifically designed for specific tasks, such as PatchTST\cite{nie2022time} and TiDE\cite{das2023long} for time series forecasting.
  
For all experimental results, we employed equitable parameter settings and they are conducted on one single NVIDIA 4090 24GB GPU. For the evaluation of time series forecasting, we utilize Mean Squared Error (MSE) and Mean Absolute Error (MAE) as primary evaluation metrics. For classification tasks, we employ Accuracy as the key indicator for assessment. \textbf{Noted that all experiments were repeated three times to report the average values}. \textbf{Bold} numbers represent the best results, and the second best are \underline{underlined}. 
Tables in Appendix present the experimental outcomes of ConvTimeNet on the tasks of time series forecasting and classification, respectively.
\subsection{Main Results Analysis}
The performance of ConvTimeNet was evaluated on both time series forecasting and classification tasks, demonstrating its robust capabilities across various datasets. For the forecasting task, the number of training epochs is fixed at 10. And we use early stopping to prevent model overfitting and set patience to 3. All other hyperparameters and initialization strategies are either derived from the authors of the original works. If no open parameter script is provided, we will set up the experimental parameters according to the default parameter settings, with a learning rate of 0.0001, batch size of 32, and dropout of 0.2.
For the classification task, the number of training epochs is fixed at 200. All the baselines reproduced in this study are implemented based on the configurations outlined in the original paper or the official code.
\begin{figure}[t]
    \centering
    \includegraphics[width=\linewidth]{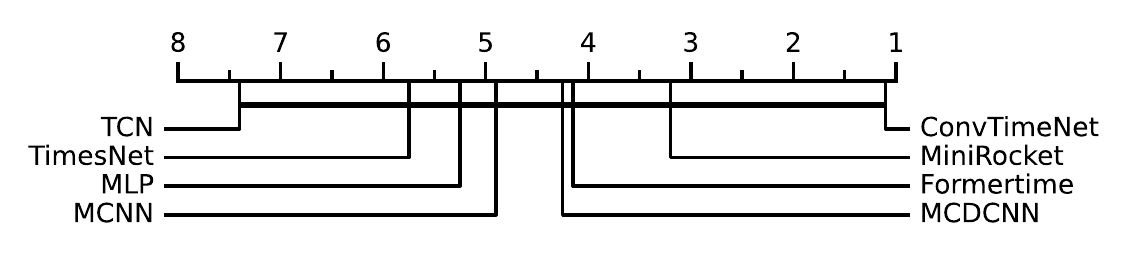}
    \caption{Critical difference diagram over the mean ranks of ConvTimeNet, baseline methods of classification task.}
    \label{fig:cd_diagram}
    \vspace{-0.15in}
\end{figure}
\begin{figure}[t]
  \centering
  \includegraphics[width=\linewidth]{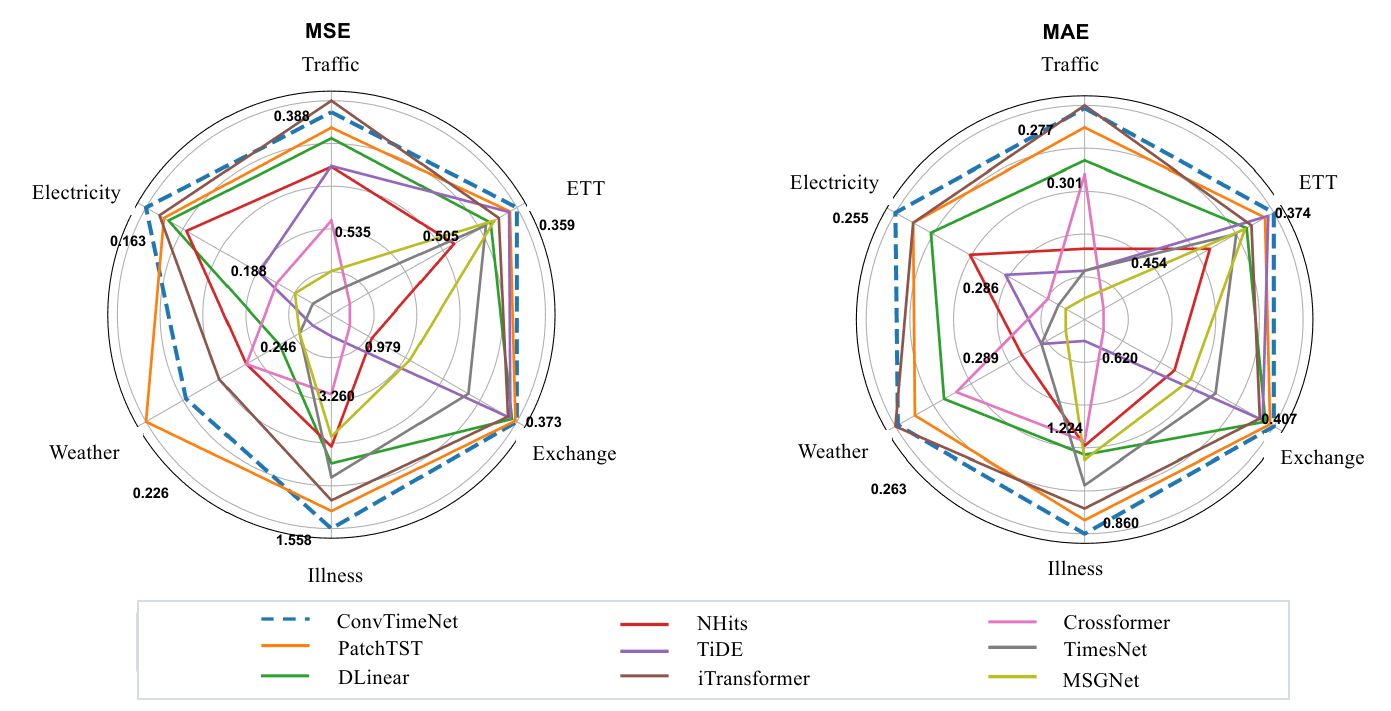}
  \caption{Model performance comparison in time series forecasting task.}
  \label{fig:radar}
  \vspace{-0.1in}
\end{figure}
\paragraph{Time Series Classification.} In classification tasks, as depicted in Figure~\ref{fig:cd_diagram} and detailed in 
Appendix B
, ConvTimeNet again outperforms current state-of-the-art methods. Figure~\ref{fig:cd_diagram} illustrates a critical difference diagram comparing the mean rankings of ConvTimeNet against various baseline methods, with rankings ranging from 1 (best) to 8 (worst). ConvTimeNet achieves the top rank, outperforming other models. The results in 
Appendix B show that ConvTimeNet surpasses other methods in over 80\% of cases across ten datasets. Compared to traditional convolutional methods like MCNN and MCDCNN, ConvTimeNet exhibits significant improvements in classification accuracy. When juxtaposed with another general-purpose network, TimesNet, ConvTimeNet shows remarkable improvements, exceeding 14\% across the ten datasets. It is worth noting that MiniRocket outperformed others on the EC and EP datasets, likely due to its varying kernel weight design, which facilitates the extraction of more refined features.
  \begin{table*}[t]
    \centering
    \caption{Experimental results w.r.t. studying the effectiveness of deformable patch embedding.}
    \vspace{-0.1in}
     \resizebox{0.95\textwidth}{!}{ \begin{tabular}{ccccccccccc}
      \toprule
        & \multicolumn{2}{c}{Pointwise} & \multicolumn{2}{c}{Uniform Patchwise} & \multicolumn{2}{c}{DePatch-Conv} & \multicolumn{2}{c}{DePatch-Conv-Conv} & \multicolumn{2}{c}{DePatch-MLP} \\
      \midrule
       Metric & Accuracy & F1 Score & Accuracy & F1 Score & Accuracy & F1 Score & Accuracy & F1 Score & Accuracy & F1 Score \\
      \midrule
      AWR & 0.961  & 0.961  & 0.979  & 0.978  & 0.981  & 0.981  & \textbf{0.987 } & \textbf{0.987 } & 0.979  & 0.978  \\
      CR & 0.958  & 0.957  & 0.982  & 0.981  & 0.965  & 0.966  & \textbf{0.986 } & \textbf{0.986 } & 0.972  & 0.972  \\
      CT & 0.968  & 0.965  & 0.995  & 0.995  & \textbf{0.996 } & \textbf{0.996 } & 0.995  & 0.995  & 0.994  & 0.994  \\
      EC & 0.311  & 0.238  & 0.328  & 0.284 & \textbf{0.356 } & \textbf{0.287 } & 0.338  & 0.249  & 0.340  & 0.253  \\
      EP & 0.986  & 0.985  & 0.978  & 0.978  & \textbf{0.990 } & \textbf{0.991 } & 0.988  & 0.988  & 0.983  & 0.983  \\
      FM & 0.550  & 0.547  & 0.663  & 0.662  & 0.677  & 0.676  & \textbf{0.680 } & \textbf{0.677 } & 0.647  & 0.645  \\
      JV & 0.984  & 0.983  & 0.987  & 0.987  & 0.985  & 0.984  & 0.990  & 0.990  & \textbf{0.992 } & \textbf{0.992 } \\
      PEMS & 0.753  & 0.746  & 0.813  & 0.806  & 0.817  & 0.809  & \textbf{0.830 } & \textbf{0.823 } & 0.821  & 0.810  \\
      SRS & 0.554  & 0.535  & 0.578  & 0.563  & 0.591  & 0.585  & \textbf{0.596 } & \textbf{0.594 } & 0.574  & 0.571  \\
      DDG & 0.527  & 0.472  & 0.540  & 0.515  & 0.560  & 0.553  & \textbf{0.660 } & \textbf{0.652 } & 0.267  & 0.160  \\
      \bottomrule
      \end{tabular}%
      }
    \label{tab:slice}%
    \vspace{-0.1in}
  \end{table*}%
\paragraph{Time Series Forecasting.} Figure~\ref{fig:radar} and 
Appendix A present a comprehensive comparison of ConvTimeNet with other models on forecasting tasks. The radar charts in Figure~\ref{fig:radar} display model performance using MSE on the left and MAE on the right, with different colored lines representing results across multiple datasets such as Traffic, ETT, and Exchange. ConvTimeNet consistently demonstrates superior performance on most datasets, particularly achieving significantly lower errors on the ILI and Weather datasets. The data in 
Appendix A further underscores these findings. ConvTimeNet delivers state-of-the-art results, outperforming in over 70\% of instances across nine distinct datasets. Notably, compared to PatchTST, ConvTimeNet shows remarkable improvements, achieving over an 8\% increase on the ETTh2 dataset and an impressive 11\% enhancement on the illness dataset. Additionally, while models like MSGNet~\cite{cai2023msgnet}—a hierarchical model with a similar structure—exhibit exceedingly high memory consumption during training on datasets like Traffic with 862 channels, ConvTimeNet maintains an optimal balance between performance and efficiency. These outcomes highlight ConvTimeNet's robust capabilities in time series forecasting tasks. However, it is important to note that ConvTimeNet does not consistently yield the best results in every predictive scenario, suggesting potential for further refinement and exploration to enhance the performance of convolutional models.
\begin{figure*}[t]
\centering
\includegraphics[width=1.0\textwidth]{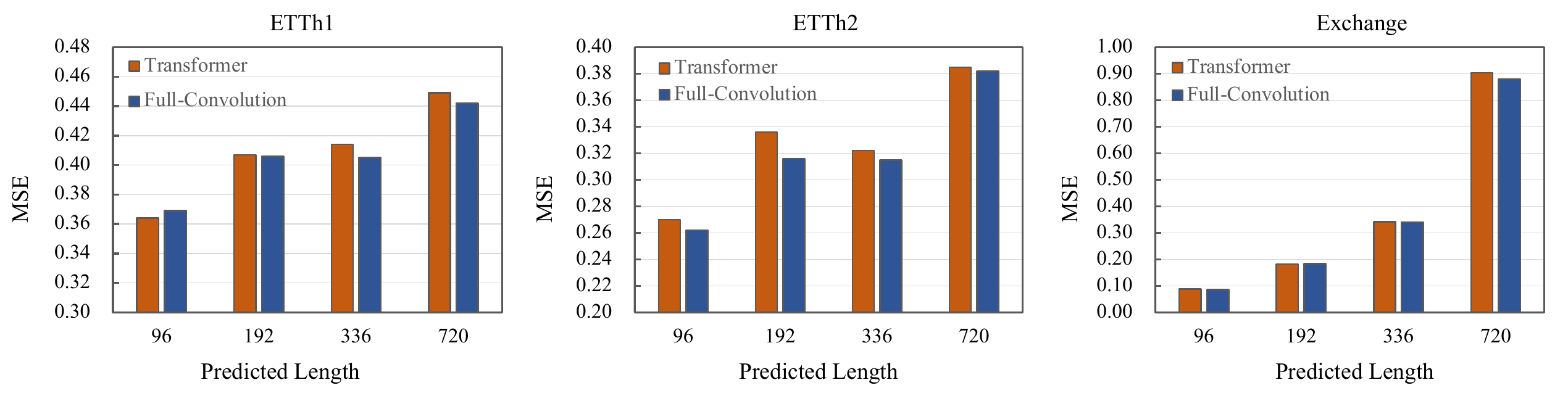}
\vspace{-0.2in}
\caption{Performance comparison between our full-convolution block and Transformer encoder block.}
\label{fig:ablation-block}
\end{figure*}
\vspace{-0.1in}
\subsection{Impacts of Deformable Patch Embedding}
The deformable patch embedding plays a vital role in the performance of ConvTimeNet, masterfully tokenizing time series data through its adaptive adjustment of the patch size and position. These adjusted patches serve as local patterns that capture the essential features of the time series. To rigorously test the effectiveness of this deformable patch mechanism, we embark on a series of experiments across ten different datasets in the realm of classification tasks. These experiments are meticulously designed to compare the outcomes of having pointwise input, implementing a uniform patch, and utilizing three distinct variants of learnable predictor networks in deformable patches. The empirical results, as detailed in Table \ref{tab:slice}, reveal a significant boost in model performance attributed to the patch operation across most datasets. Particularly noteworthy is its impact on the FM dataset, which records electroencephalogram signals under different finger movements, where it catalyzed a remarkable 20\% leap in performance. The prowess of the deformable patch is further highlighted in specific cases, such as the DDG dataset, recording audio information of different animals, where employing a deformable patch configured with a two-layer convolutional network led to a substantial 22\% enhancement in model performance. Among the various deformable patch configurations tested, the variant with a two-layer convolutional network emerged as the most effective. In addition, three datasets achieve the best results under the single layer convolution setting, which indicates the deformable patch can be equipped with different predictors in various scenarios. At the same time, the experimental results validate that convolutional neural networks in our model are capable of performing local pattern extraction.

\subsection{Effectiveness of Fully Convolutional Block}
To validate the effectiveness of our designed fully convolutional block, we verify the effectiveness of the block on the forecasting task. Due to space limitations, we chose three datasets for the experiment. We replace the fully convolutional block with a traditional transformer encoder block and retain the learnable residual, where the building blocks of our model comprise a minimum of six layers. It is noteworthy that under the condition of setting the transform block to six layers, the model does not exhibit obvious overfitting phenomena, possibly due to the effect of the learnable residual. This allows for a direct comparison of their forecasting capabilities, as depicted in Figure~\ref{fig:ablation-block}. In a majority of cases, the fully convolutional block demonstrates greater performance than the transformer encoder block. This effectively validates the superiority of the fully convolutional block in time series forecasting tasks. However, on the Exchange dataset, with about 5000 exchange rate data points, the fully convolution block just slightly surpasses the transformer encoder block. This might be attributed to the smaller sample size of the Exchange dataset, thereby causing both blocks to receive insufficient training.
\subsection{Impacts of Deep Hierarchical Architecture}
 \begin{table*}[t]
    \centering
    \vspace{-0.1in}
      \caption{Experimental results w.r.t. studying the effectiveness of deep hierarchical architecture.}
      \vspace{-0.1in}
      \resizebox{\textwidth}{!} {    
      \begin{tabular}{c|cccccccc}
      \toprule
      \multicolumn{1}{c}{Models} & \multicolumn{3}{c}{One Stage} & \multicolumn{3}{c}{Two Stages} & Three Stages \\
      \midrule
       Setting & [7,7,7,7,7,7] & [13,13,13,13,13,13] & [19,19,19,19,19,19] & [7,7,7,13,13,13] & [7,7,7,19,19,19] & [13,13,13,19,19,19] & [7,7,13,13,19,19]  \\
       \hline
        CT & 0.992 & 0.994 & \textbf{0.995} & \textbf{0.995} & 0.993 & \textbf{0.995} & \textbf{0.995} \\
        FM & 0.640 & \textbf{0.680} & 0.653 & 0.643 & 0.660 & 0.630 & \textbf{0.680} \\
        JV & 0.990 & 0.990 & 0.988 & 0.987 & 0.991 & \textbf{0.992} & 0.990 \\
        \hline
        Setting & [19,19,19,19,19,19] & [29,29,29,29,29,29] & [37,37,37,37,37,37] & [19,19,19,29,29,29] & [19,19,19,37,37,37] & [29,29,29,37,37,.37] & [19,19,29,29,37,37]  \\
             \hline
  
        CR & 0.982 & 0.977 & 0.982 & 0.977 & 0.958 & 0.972 & \textbf{0.986} \\
        AWR & 0.984 & 0.982 & 0.975 & 0.982 & 0.980 & 0.979 & \textbf{0.987} \\
        EC & \textbf{0.347} & 0.345 & 0.342 & 0.338 & 0.340 & 0.337 & 0.338 \\
        PEMS & 0.803 & 0.807 & 0.829 & \textbf{0.847} & 0.819 & 0.807 & 0.830 \\
        DDG & 0.573 & 0.560 & 0.500 & 0.580 & 0.513 & 0.513 & \textbf{0.660} \\
        \hline
        Setting & [37,37,37,37,37,37] & [43,43,43,43,43,43] & [53,53,53,53,53,53] & [37,37,37,43,43,43] & [37,37,37,53,53,53] & [43,43,43,53,53,53] & [37,37,43,43,53,53] \\
             \hline
        EP & \textbf{0.988} & 0.983 & 0.981 & 0.980 & 0.978 & 0.976 & \textbf{0.988} \\
        SRS2 & 0.585 & 0.570 & 0.589 & 0.580 & 0.582 & 0.589 & \textbf{0.596} \\
      \midrule
      \end{tabular}%
      }
    \label{tab:hiearachy}%
  \end{table*}%
To gain a comprehensive understanding of how varying hierarchical structures influence model performance, we conduct an extensive series of experiments. These experiments are conducted across ten datasets dedicated to classification tasks, utilizing seven distinct hierarchical configurations. The objective is to determine the impact of hierarchical depth and complexity on model efficacy. The results, as tabulated in Table \ref{tab:hiearachy}, reveal a consistent trend: models configured with a three-stage hierarchical structure mostly outperform those with simpler one-stage and two-stage structures. This superior performance underscores the value of a more complex hierarchical arrangement in handling classification tasks. Notably, the experiment results demonstrate that an incremental kernel size approach to hierarchy, where each stage employs a larger kernel than the previous, tends to yield more favorable outcomes in most scenarios. One plausible explanation for this observation is that as the model delves deeper and the building block kernel becomes larger, it progressively makes full use of multi-scale information. This revision emphasizes that with increased depth and larger kernels, the model enhances the richness of interaction between representations at different scales of the temporal basic patterns.
\vspace{-0.2in}
\subsection{The Ablation of Learnable Residual}
To mitigate the risk of overfitting potentially associated with the extensive stacking of fully convolutional blocks, we introduced the concept of learnable residual within the architecture of ConvTimeNet. This feature is specifically utilized to balance the model's complexity with its learning capacity. To empirically evaluate the impact of learnable residual, we conduct a series of ablation studies across ten distinct datasets, focusing on classification tasks.
The results of these studies are illuminating in Table~\ref{tab:ablation_lresidual}. On average, the incorporation of learnable residual into ConvTimeNet led to a 4\% enhancement in model performance. This improvement can likely be attributed to the learnable residual, which can dynamically adjust the contribution of each layer to the final output, enabling the model to effectively capture and represent more complex patterns in the data without succumbing to overfitting.
\begin{table}[t]
  \centering
  \caption{Experimental results w.r.t. studying the effectiveness of learnable residual.}
  \vspace{-0.1in}
    \resizebox{0.95\linewidth}{!}{\begin{tabular}{ccccc}
    \toprule
    \multirow{1}[2]{*}{} & \multicolumn{2}{c}{W/ Learnable Redsidual} & \multicolumn{2}{c}{W/O Learnable Residual} \\
    \midrule
     Metric & Accuracy & F1 Score & Accuracy & F1 Score \\
    \midrule
    AWR & \textbf{0.987 } & \textbf{0.987 } & 0.986  & 0.986  \\
    CR & \textbf{0.986 } & \textbf{0.986 } & 0.949  & 0.948  \\
    CT & \textbf{0.995 } & \textbf{0.995 } & \textbf{0.995}  & \textbf{0.995}  \\
    EC & \textbf{0.338 } & 0.249 & 0.331  & \textbf{0.266}  \\
    EP & \textbf{0.988 } & \textbf{0.988 } & 0.978  & 0.979  \\
    FM & \textbf{0.680 } & \textbf{0.677 } & 0.627  & 0.626  \\
    JV & \textbf{0.990 } & \textbf{0.990 } & 0.989  & \textbf{0.990}  \\
    PEMS & 0.830  & 0.823  & \textbf{0.873 } & \textbf{0.870 } \\
    SRS & 0.596  & 0.594  & \textbf{0.624 } & \textbf{0.621 } \\
    DDG & \textbf{0.660 } & \textbf{0.652 } & 0.467  & 0.415  \\
    \bottomrule
    \end{tabular}%
    \vspace{-0.3in}
    }
  \label{tab:ablation_lresidual}%
\end{table}%
\begin{figure*}[t]
  \centering
    \caption{The visualization of the forecasting case, generated by various models under the input-336-predict-336 setting, is presented. The black lines represent the ground truth, while the orange lines represent the predicted values.}
  \includegraphics[width=1.0\textwidth]{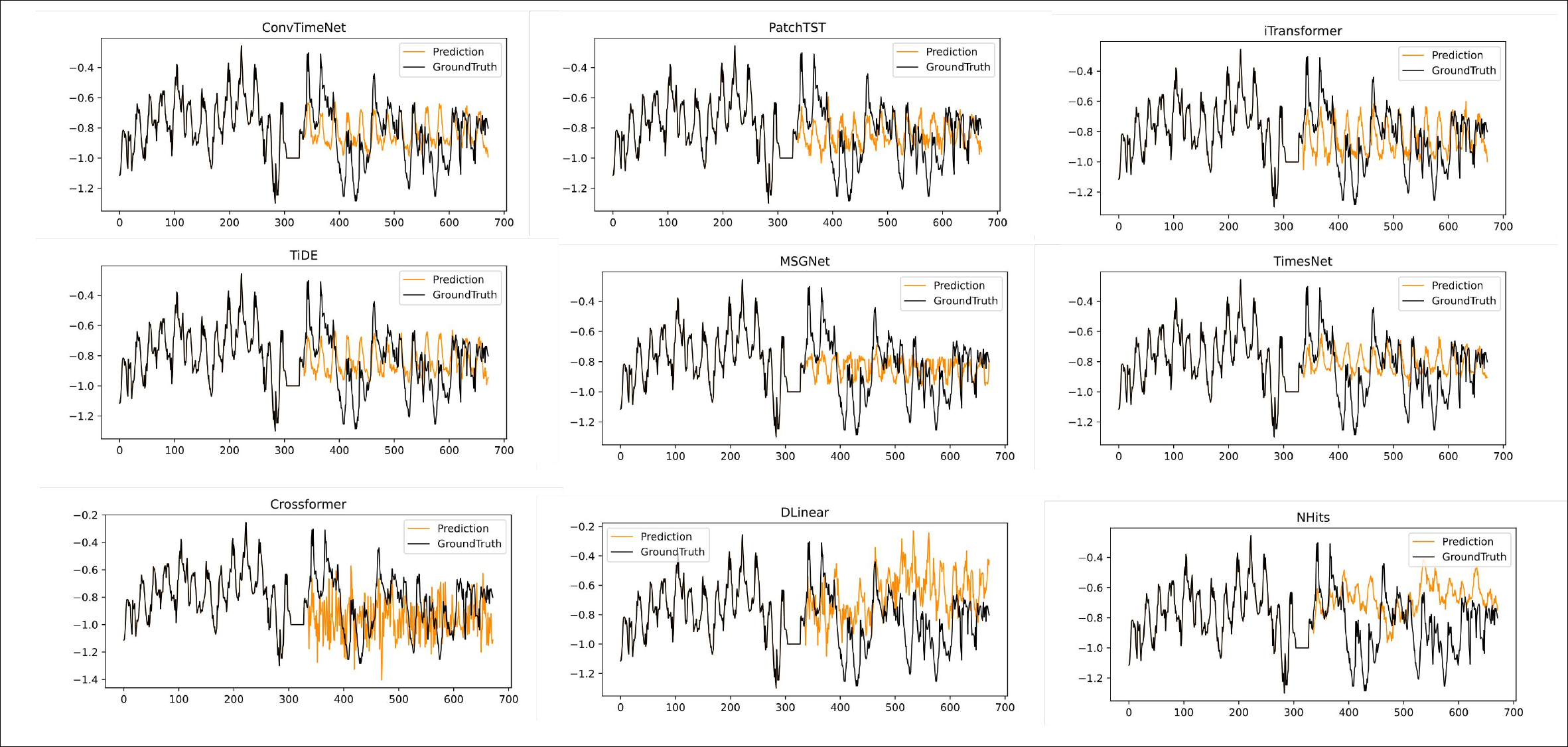}
  \label{fig:prediction_case}
  \vspace{-0.1in}
\end{figure*}

\subsection{Case Study}
We present visual cases for time series forecasting in Figure~\ref{fig:prediction_case}. In each subplot, the black line represents the ground truth values, while the orange line shows the predicted values produced by the corresponding model. For this analysis, we chose DLinear, NHiTS, and TiDE as representatives of MLP-based models. PatchTST, iTransformer, and CrossFormer as representatives of Transformer-based models and MSGNet and TimesNet as representatives of convolutional models similar to ours. These visual representations provide a clear comparison among various models. The study randomly selects a sample from the test dataset as input and plots the predicted results in all models. It can be intuitively observed from the chart that the predicted trend of ConvTimeNet is more stable compared to DLinear~\cite{zeng2023transformers} and NHits\cite{challu2023nhits}. Although MSGNet~\cite{cai2023msgnet} also shows relative stability, the predicted trend of ConvTimeNet is more in line with ground truth.

\subsection{Limitation Discussions}
While our proposed ConvTimeNet generally achieves superior performance and reduced computational burden, it is essential to acknowledge its limitations in two key areas. Firstly, there is potential for further enhancement in capturing cross-channel dependencies in multivariate time series tasks. In scenarios where multiple variables interact over time, comprehending the dependencies and correlations between different channels becomes crucial for accurate modeling and prediction. By considering cross-channel interactions starting from the local pattern extraction stage, the model could be refined to more effectively capture the intricate relationships present in multivariate data. Designing an extraction module that holistically integrates local features across multiple variables would provide a more comprehensive understanding of the data, potentially leading to improved performance in tasks requiring nuanced inter-variable analysis. Secondly, for certain datasets, we observed that the hierarchical hyperparameters of ConvTimeNet necessitate meticulous tuning specific to the dataset in question. This process is not only time-consuming but also adds to the implementation costs. To address this challenge, future work could explore the application of neural architecture search~\cite{tan2019efficientnet} to facilitate the automatic tuning of hyperparameters, thereby alleviating this burden. Thirdly, our study only validated the effectiveness of ConvTimeNet within the supervised learning paradigm without extending experiments to the realm of transfer learning. Recent studies~\cite{cheng2023timemae} have demonstrated that self-supervised pre-training can significantly enhance the performance of deep learning architectures. Consequently, our approach might not fully realize its potential without the benefits of self-supervised pre-training. Future research could consider integrating self-supervised learning strategies with ConvTimeNet to further enhance its applicability and effectiveness.

\section{Conclusion}
In this study, we delved into the two major challenges faced in time series data analysis: accurate local pattern extraction and effective global sequence dependency capturing. We addressed the research question of how to reinvigorate the role of convolutional networks in time series modeling. Specifically, we proposed ConvTimeNet, a deep hierarchical fully convolutional network that serves as a versatile backbone for time series analysis. A key finding of our work is that maintaining a deep and hierarchical convolutional architecture, equipped with modern techniques, can yield superior or competitive performance compared to prevalent Transformer networks and pioneering convolutional models. Extensive experiments conducted on time series forecasting and classification tasks fully substantiate its effectiveness. Overall, we hope that ConvTimeNet can serve as an alternative model and encourage the research community to rethink the importance of convolution in time series mining tasks. By effectively unifying local pattern extraction and global dependency modeling within a single framework, ConvTimeNet demonstrates that convolutional networks still hold significant potential for advancing time series analysis.
\clearpage
\bibliographystyle{ACM-Reference-Format}
\bibliography{sample-base}
\clearpage
\appendix
 \begin{table*}[t]
  \small
   \renewcommand\arraystretch{1.0}
   \tabcolsep=0.12cm
   \centering
   \caption*{Appendix A: Experimental results of time series forecasting task evaluated by MSE and MAE.}
   \vspace{-0.1in}
     \resizebox{0.9\textwidth}{!}{
     \begin{tabular}{c|c|cc|cccccccccccccccc}
     \toprule
     \multicolumn{2}{c|}{\multirow{1}{*}{Methods}} & \multicolumn{2}{c|}{\multirow{1}{*}{ConvTimeNet}} & \multicolumn{2}{c}{\multirow{1}{*}{PatchTST}} & \multicolumn{2}{c}{\multirow{1}{*}{iTransformer}} & \multicolumn{2}{c}{\multirow{1}{*}{TiDE}} & \multicolumn{2}{c}{\multirow{1}{*}{MSGNet}} & \multicolumn{2}{c}{\multirow{1}{*}{TimesNet}} & \multicolumn{2}{c}{\multirow{1}{*}{Crossformer}} & \multicolumn{2}{c}{\multirow{1}{*}{DLinear}} & \multicolumn{2}{c}{\multirow{1}{*}{NHits}} \\
     \midrule
     \multicolumn{2}{c|}{Metric} & MSE & MAE & MSE & \multicolumn{1}{c}{MAE} & MSE & \multicolumn{1}{c}{MAE} & MSE & \multicolumn{1}{c}{MAE} & MSE & \multicolumn{1}{c}{MAE} & MSE & \multicolumn{1}{c}{MAE} & MSE & \multicolumn{1}{c}{MAE} & MSE & \multicolumn{1}{c}{MAE} & MSE & \multicolumn{1}{c}{MAE} \\
     \midrule
     \multirow{4}[2]{*}{\begin{sideways}ETTh1\end{sideways}} & 96 & \textbf{0.368 } & \textbf{0.394 } & 0.385  & 0.408  & 0.405  & 0.419  & \underline{0.374}  & \underline{0.395}  & 0.423  & 0.440  & \textcolor[rgb]{ .122,  .137,  .161}{0.421 } & \textcolor[rgb]{ .122,  .137,  .161}{0.438 } & 0.390  & 0.417  & 0.375  & 0.396  & 0.423  & 0.444  \\
       & 192 & \textbf{0.406 } & \textbf{0.414 } & 0.419  & 0.426  & 0.448  & 0.447  & \underline{0.409}  & \underline{0.417}  & 0.465  & 0.469  & \textcolor[rgb]{ .122,  .137,  .161}{0.482 } & \textcolor[rgb]{ .122,  .137,  .161}{0.479 } & 0.424  & 0.448  & 0.428  & 0.437  & 0.504  & 0.493  \\
       & 336 & \textbf{0.405 } & \textbf{0.420 } & \underline{0.429}  & 0.434  & 0.482  & 0.470  & 0.435  &\underline{0.433}  & 0.468  & 0.473  & \textcolor[rgb]{ .122,  .137,  .161}{0.528 } & \textcolor[rgb]{ .122,  .137,  .161}{0.505 } & 0.486  & 0.492  & 0.448  & 0.449  & 0.513  & 0.503  \\
       & 720 & \textbf{0.442 } & \textbf{0.457 } & \underline{0.446}  & 0.462  & 0.560  & 0.537  & \underline{0.446}  & \underline{0.460}  & 0.540  & 0.524  & \textcolor[rgb]{ .122,  .137,  .161}{0.527 } & \textcolor[rgb]{ .122,  .137,  .161}{0.510 } & 0.507  & 0.519  & 0.505  & 0.514  & 0.622  & 0.564  \\
     \midrule
     \multirow{4}[2]{*}{\begin{sideways}ETTh2\end{sideways}} & 96 & \textbf{0.264 } & \textbf{0.330 } & \underline{0.278}  & \underline{0.341}  & 0.305  & 0.361  & 0.290  & 0.350  & 0.348  & 0.399  & \textcolor[rgb]{ .122,  .137,  .161}{0.355 } & \textcolor[rgb]{ .122,  .137,  .161}{0.408 } & 0.803  & 0.628  & 0.296  & 0.360  & 0.318  & 0.373  \\
       & 192 & \textbf{0.316 } & \textbf{0.368 } & \underline{0.343}  & \underline{0.382}  & 0.391  & 0.412  & 0.349  & 0.388  & 0.404  & 0.431  & \textcolor[rgb]{ .122,  .137,  .161}{0.403 } & \textcolor[rgb]{ .122,  .137,  .161}{0.434 } & 1.028  & 0.743  & 0.391  & 0.423  & 0.425  & 0.447  \\
       & 336 & \textbf{0.315 } & \textbf{0.378 } & 0.372  & \underline{0.404}  & 0.418  & 0.433  & \underline{0.371}  & 0.409  & 0.375  & 0.419  & \textcolor[rgb]{ .122,  .137,  .161}{0.398 } & \textcolor[rgb]{ .122,  .137,  .161}{0.434 } & 1.167  & 0.828  & 0.445  & 0.460  & 0.596  & 0.527  \\
       & 720 & \textbf{0.382 } & \textbf{0.425 } & \underline{0.395}  & \underline{0.430}  & 0.437  & 0.455  & 0.401  & 0.437  & 0.421  & 0.451  & \textcolor[rgb]{ .122,  .137,  .161}{0.443 } & \textcolor[rgb]{ .122,  .137,  .161}{0.465 } & 1.665  & 1.032  & 0.700  & 0.592  & 1.353  & 0.810  \\
     \midrule
     \multirow{4}[2]{*}{\begin{sideways}ETTm1\end{sideways}}& 96 & \textbf{0.292 } & \textbf{0.345 } & \underline{0.298}  & \textbf{0.345}  & 0.306  & 0.360  & 0.310  & 0.352  & 0.309  & 0.362  & \textcolor[rgb]{ .122,  .137,  .161}{0.331 } & \textcolor[rgb]{ .122,  .137,  .161}{0.372 } & 0.345  & 0.394  & 0.303  & \underline{0.346}  & 0.323  & 0.376  \\
       & 192 & \textbf{0.329 } & \textbf{0.368 } & 0.339  & 0.374  & 0.345  & 0.382  & 0.345  & \underline{0.372}  & 0.356  & 0.392  & \textcolor[rgb]{ .122,  .137,  .161}{0.435 } & \textcolor[rgb]{ .122,  .137,  .161}{0.421 } & 0.461  & 0.483  & \underline{0.338}  & \textbf{0.368 } & 0.365  & 0.401  \\
       & 336 & \textbf{0.363 } & \underline{0.390}  & 0.381  & 0.401  & 0.378  & 0.402  & 0.379  & 0.391  & 0.393  & 0.414  & \textcolor[rgb]{ .122,  .137,  .161}{0.457·} & \textcolor[rgb]{ .122,  .137,  .161}{0.445 } & 0.623  & 0.586  & \underline{0.373}  & \textbf{0.389 } & 0.400  & 0.423  \\
       & 720 & \textbf{0.427 } & \underline{0.428}  & \underline{0.428}  & 0.431  & 0.443  & 0.439  & 0.435  & 0.423  & 0.440  & 0.445  & \textcolor[rgb]{ .122,  .137,  .161}{0.526 } & \textcolor[rgb]{ .122,  .137,  .161}{0.481 } & 0.673  & 0.593  & \underline{0.428}  & \textbf{0.423 } & 0.463  & 0.463  \\
     \midrule
     \multirow{4}[2]{*}{\begin{sideways}ETTm2\end{sideways}} & 96 & \textbf{0.167 } & \textbf{0.257 } & 0.174  & \underline{0.261}  & 0.174  & 0.266  & \textbf{0.167 } & \textbf{0.257 } & 0.188  & 0.273  & \textcolor[rgb]{ .122,  .137,  .161}{0.190 } & \textcolor[rgb]{ .122,  .137,  .161}{0.276 } & 0.330  & 0.401  & \underline{0.170}  & 0.264  & 0.212  & 0.290  \\
       & 192 & \textbf{0.222 } & \underline{0.295}  & 0.238  & 0.307  & 0.247  & 0.315  & \underline{0.223}  & \textbf{0.294 } & 0.246  & 0.315  & \textcolor[rgb]{ .122,  .137,  .161}{0.244 } & \textcolor[rgb]{ .122,  .137,  .161}{0.311 } & 0.623  & 0.543  & 0.233  & 0.311  & 0.270  & 0.330  \\
       & 336 & \textbf{0.276 } & \textbf{0.329 } & 0.293  & 0.346  & 0.292  & \underline{0.343}  &\underline{0.277}  & \textbf{0.329 } & 0.301  & 0.347  & \textcolor[rgb]{ .122,  .137,  .161}{0.302 } & \textcolor[rgb]{ .122,  .137,  .161}{0.349 } & 0.887  & 0.637  & 0.298  & 0.358  & 0.340  & 0.377  \\
       & 720 & \textbf{0.358 } & \textbf{0.381 } & 0.373  & 0.401  & 0.375  & 0.395  & \underline{0.371}  & \underline{0.386}  & 0.407  & 0.411  & \textcolor[rgb]{ .122,  .137,  .161}{0.406 } & \textcolor[rgb]{ .122,  .137,  .161}{0.406 } & 0.844  & 0.640  & 0.423  & 0.437  & 0.444  & 0.444  \\
     \midrule
     \multirow{4}[2]{*}{\textcolor[rgb]{ .122,  .137,  .161}{\begin{sideways}Electricity\end{sideways}}} & 96 & \textbf{0.132 } & \textbf{0.226 } & \underline{0.138}  & 0.233  & \textbf{0.132 } & \underline{0.228}  & 0.160  & 0.262  & 0.169  & 0.279  & 0.177  & 0.281  & 0.150  & 0.258  & 0.141  & 0.238  & 0.147  & 0.250  \\
       & 192 & \textbf{0.148 } & \textbf{0.241 } & \underline{0.153}  & \underline{0.247}  & 0.154  & 0.249  & 0.174  & 0.275  & 0.188  & 0.296  & 0.193  & 0.295  & 0.175  & 0.284  & 0.154  & 0.251  & 0.154  & 0.261  \\
       & 336 & \textbf{0.165 } & \textbf{0.259 } & \underline{0.170}  & \underline{0.263}  & 0.172  & 0.267  &  0.190 & 0.289  & 0.199  & 0.307  & 0.206  & 0.306  & 0.218  & 0.325  & 0.170  & 0.269  & 0.177  & 0.280  \\
       & 720 & \underline{0.205}  & \textbf{0.293 } & 0.206  & \underline{0.295}  & \textbf{0.204 } & 0.296  & 0.229  & 0.319  & 0.227  & 0.330  & 0.223  & 0.320  & 0.226  & 0.324  & 0.205  & 0.302  & 0.211  & 0.313  \\
     \midrule
     \multirow{4}[2]{*}{\begin{sideways}Exchange\end{sideways}} & 96 & \textbf{0.086 } & \textbf{0.204 } & 0.094  & 0.216  & 0.099  & 0.225  & 0.107  & 0.233  & 0.167  & 0.303  & 0.166  & 0.305  & 0.283  & 0.393  & \underline{0.087}  & \underline{0.214}  & 0.126  & 0.256  \\
       & 192 & \underline{0.184}  & \underline{0.303}  & 0.191  & 0.311  & 0.206  & 0.329  & 0.201  & 0.323  & 0.302  & 0.402  & 0.303  & 0.413  & 1.087  & 0.804  & \textbf{0.164 } & \textbf{0.298 } & 0.381  & 0.455  \\
       & 336 & \underline{0.341}  & \textbf{0.421 } & 0.343  & \underline{0.427}  & 0.370  & 0.448  & 0.351  & 0.432  & 0.581  & 0.556  & 0.445  & 0.511  & 1.367  & 0.905  & \textbf{0.333 } & 0.437  & 0.576  & 0.577  \\
       & 720 & \textbf{0.879 } & \textbf{0.701 } & \underline{0.888}  & \underline{0.706}  & 0.963  & 0.746  & 0.940  & 0.735  & 2.229  & 1.079  & 1.389  & 0.899  & 1.546  & 0.987  & 0.988  & 0.749  & 2.831  & 1.192  \\
     \midrule
     \multirow{4}[2]{*}{\begin{sideways}Traffic\end{sideways}} & 96 & \underline{0.376}  & \textbf{0.265 } & 0.395  & 0.272  & \textbf{0.361 } & \underline{0.266}  &  0.445 & 0.325  & 0.567  & 0.337  & 0.600  & 0.323  & 0.514  & 0.292  & 0.411  & 0.284  & 0.436  & 0.313  \\
       & 192 & \underline{0.392}  & \textbf{0.271 } & 0.411  & \underline{0.278}  & \textbf{0.378 } & \textbf{0.271 } & 0.458  & 0.331  & 0.579  & 0.339  & 0.612  & 0.327  & 0.528  & 0.298  & 0.423  & 0.289  & 0.449  & 0.317  \\
       & 336 & \underline{0.405}  & \underline{0.277}  & 0.424  & 0.284  & \textbf{0.390 } & \textbf{0.274 } &  0.471 & 0.336  & 0.604  & 0.350  & 0.628  & 0.344  & 0.538  & 0.303  & 0.437  & 0.297  & 0.475  & 0.330  \\
       & 720 & \underline{0.436}  & \underline{0.294}  & 0.453  & 0.300  & \textbf{0.424 } & \textbf{0.291 } & 0.500 &  0.352  & 0.637  & 0.359  & 0.657  & 0.349  & 0.559  & 0.312  & 0.467  & 0.316  & 0.517  & 0.350  \\
     \midrule
     \multirow{4}[2]{*}{\begin{sideways}Weather \end{sideways}} & 96& 0.155  & \underline{0.205}  & \textbf{0.147 } & \textbf{0.197 } & 0.162  & 0.212  & 0.178  & 0.229  & 0.160  & 0.217  & 0.168  & 0.225  & \underline{0.148}  & 0.214  & 0.176  & 0.236  & 0.156  & 0.212  \\
       & 192 & \underline{0.200}  & \underline{0.249}  & \textbf{0.191 } & \textbf{0.240 } & 0.205  & 0.251  & 0.221  & 0.264  & 0.208  & 0.255  & 0.218  & 0.268  & 0.201  & 0.270  & 0.217  & 0.275  & 0.203  & 0.260  \\
       & 336 & 0.252  & \underline{0.287}  & \textbf{0.244 } & \textbf{0.282 } & 0.257  & 0.291  & 0.268  & 0.298  & 0.272  & 0.302  & 0.269  & 0.301  & \underline{0.248}  & 0.311  & 0.264  & 0.315  & 0.266  & 0.319  \\
       & 720 & \underline{0.321}  & \underline{0.335}  & \textbf{0.320 } & \textbf{0.334 } & 0.325  & 0.337  & 0.336  & 0.345  & 0.357  & 0.356  & 0.340  & 0.350  & 0.366  & 0.395  & 0.325  & 0.364  & 0.338  & 0.365  \\
     \midrule
     \multirow{4}[2]{*}{\begin{sideways}Illness \end{sideways}} & 24 & \textbf{1.469 } & \textbf{0.800 } & \underline{1.657}  & \underline{0.869}  & 1.930  & 0.883  & 3.988  & 1.444  & 2.453  & 1.044  & 2.130  & 0.981  & 4.312  & 1.418  & 2.313  & 1.059  & 2.575  & 1.146  \\
       & 36 & \textbf{1.450 } &\underline{0.845}  & \underline{1.467}  & \textbf{0.813 } & 1.807  & 0.904  & 3.937  & 1.432  & 2.628  & 1.021  & 2.312  & 1.013  & 4.034  & 1.319  & 2.402  & 1.102  & 2.676  & 1.128  \\
       & 48 & \textbf{1.572 } & \textbf{0.875 } & \underline{1.833}  & \underline{0.913}  & 1.894  & 0.945  & 4.074  & 1.451  & 2.808  & 1.106  & 2.334  & 1.042  & 4.303  & 1.393  & 2.420  & 1.111  & 2.607  & 1.135  \\
       & 60 & \textbf{1.741 } & \textbf{0.920 } & 2.168  & 1.009  & \underline{2.033}  & 1.013  & 3.981  & 1.434  & 3.008  & 1.157  & 2.051  & \underline{0.989}  & 0.390  & 0.417  & 2.400  & 1.116  & 2.531  & 1.087  \\
     \bottomrule
     \end{tabular}%
     }
   \label{tab:forecast-main}%
 \end{table*}

 \begin{table*}[t]
  \tiny
   \renewcommand\arraystretch{1.0}
  \tabcolsep=0.12cm
   \centering
   \caption*{Appendix B: Results of classification task evaluated by Accuracy. $-$ indicates the model couldn't run due to insufficient memory.}
   \vspace{-0.1in}
     \resizebox{0.9\textwidth}{!}{
     \begin{tabular}{c|c|ccccccccc}
     \toprule
      \multicolumn{1}{c}{\multirow{1}{*}{Datasets}} & \multicolumn{1}{c}{\multirow{1}{*}{ConvTimeNet}} & \multicolumn{1}{c}{\multirow{1}{*}{FormerTime}} & \multicolumn{1}{c}{\multirow{1}{*}{TimesNet}} & \multicolumn{1}{c}{\multirow{1}{*}{MiniRocket}} & \multicolumn{1}{c}{\multirow{1}{*}{TST}} & \multicolumn{1}{c}{\multirow{1}{*}{MLP}} & \multicolumn{1}{c}{\multirow{1}{*}{TCN}} & \multicolumn{1}{c}{\multirow{1}{*}{MCNN}} & \multicolumn{1}{c}{\multirow{1}{*}{MCDCNN}} \\
     \midrule
     AWR & \textbf{0.987 } & 0.978  & \underline{0.980}  & 0.972  & \underline{0.980}  & 0.960  & 0.884  & 0.977  & \underline{0.980}  \\
       CR & \textbf{0.986 } & 0.917  & 0.889  & \underline{0.981}  & 0.898  & 0.935  & 0.868  & 0.917  & 0.870  \\
       CT & \textbf{0.995 } & \underline{0.992}  & 0.984  & 0.987  & 0.990  & 0.955  & 0.968  & 0.991  & 0.988  \\
       EC & \textbf{0.338 } & 0.312  & 0.287  & \underline{0.327}  & - & 0.309  & 0.299  & 0.293  & 0.298  \\
       EP & \underline{0.988}  & 0.952  & 0.902  & \textbf{0.993 } & 0.901  & 0.961  & 0.942  & 0.961  & 0.971  \\
       FM & \textbf{0.680 } & \underline{0.650}  & 0.610  & 0.638  & 0.580  & 0.607  & 0.580  & 0.597  & 0.617  \\
       JV & \textbf{0.990 } & 0.986  & 0.981  & \underline{0.987}  & 0.986  & 0.983  & 0.974  & 0.977  & 0.978  \\
       PEMS & \textbf{0.830 } & 0.173  & 0.728  & 0.795  & 0.778  & 0.761  & 0.669  & \underline{0.804}  & 0.792  \\
       SRS & \textbf{0.596 } & 0.572  & 0.542  & 0.564  & 0.541  & 0.546  & 0.507  & \underline{0.574}  & 0.565  \\
       DDG & \textbf{0.660 } & 0.240  & 0.400  & 0.620  & 0.340  & 0.280  & 0.200  & 0.213  & \underline{0.624}  \\				
     \bottomrule
     \end{tabular} %
     }
   \label{tab:classification-main}%
 \end{table*}%

\end{document}